\documentclass[runningheads]{llncs}

\usepackage{eccv}

\usepackage[pagebackref,breaklinks,colorlinks]{hyperref}
\usepackage{eccvabbrv}
\usepackage[accsupp]{axessibility}  % Improves PDF readability for those with disabilities.
\usepackage{url}            % simple URL typesetting
\usepackage{booktabs}       % professional-quality tables
\usepackage{amsfonts}       % blackboard math symbols
\usepackage{nicefrac}       % compact symbols for 1/2, etc.
\usepackage{xcolor}       % colors
\usepackage{xkcdcolors}
\usepackage{epsfig}
\usepackage{graphicx}
\usepackage{placeins}
\usepackage{multirow}
\usepackage[skip=5pt]{caption}
\usepackage{rotating}
\usepackage{cancel}
\usepackage{setspace}
\usepackage{eso-pic}

\usepackage{bm}
\usepackage{bibunits} % for separate ref list in appendix

\usepackage{amssymb}
\usepackage{amsmath}
\usepackage{graphicx}
\usepackage{wrapfig,lipsum,booktabs}

\usepackage{epsfig}
\usepackage{tikz}
\usepackage{algpseudocode}
\usepackage{algorithm}
\usepackage{mathrsfs}

\usepackage{booktabs}       % professional-quality tables

\usepackage{thmtools,thm-restate}
\usepackage[capitalize,noabbrev]{cleveref}
\usepackage{listings}
\usepackage{lstautogobble}  %
\usepackage{color}          %
\usepackage{colortbl}
\usepackage{zi4}            %

\usepackage{subcaption}        % subfigure
\usepackage{siunitx}               % <--- new
\usepackage[export]{adjustbox}
\usepackage{makecell}
\usepackage{url}
\usepackage{tikz, pgfplots}
\usetikzlibrary{positioning}
\usepackage{capt-of}
\usepackage{diagbox}

\usepackage{orcidlink}
\usepackage{slashbox}
\DeclareMathOperator*{\argmin}{arg\,min}

\newcommand{\yb}{{\boldsymbol y}}

\newcommand{\nb}{{\boldsymbol n}}

\newcommand{\s}{{\boldsymbol s}}

\newcommand{\x}{{\boldsymbol x}}
\newcommand{\y}{{\boldsymbol y}}
\newcommand{\z}{{\boldsymbol z}}

\newcommand{\epsilonb}{{\boldsymbol \epsilon}}

\newcommand{\X}{{\boldsymbol X}}
\newcommand{\Y}{{\boldsymbol Y}}

\newcommand{\Ab}{{\boldsymbol A}}

\newcommand{\Tb}{{\boldsymbol T}}
\newcommand{\Ib}{{\boldsymbol I}}

\newcommand{\Ed}{{\mathbb E}}
\newcommand{\Rd}{{\mathbb R}}

\newcommand{\Nc}{{\mathcal N}}

\newcommand{\Oc}{{\mathcal O}}

\newcommand{\Lc}{{\mathcal L}}

\newcommand{\code}[1] {\texttt{#1}}

\definecolor{trolleygrey}{rgb}{0.5, 0.5, 0.5}
\definecolor{BrickRed}{rgb}{0.6,0,0}
\definecolor{RoyalBlue}{rgb}{0,0,0.8}
\definecolor{Tdgreen}{rgb}{0,0.4,0.7}
\definecolor{pinegreen}{rgb}{0.0, 0.47, 0.44}
\definecolor{cornellred}{rgb}{0.7, 0.11, 0.11}
\definecolor{cadmiumgreen}{rgb}{0.0, 0.42, 0.24}
\definecolor{spirodiscoball}{rgb}{0.06, 0.75, 0.99}
\definecolor{mylightblue}{rgb}{0.85, 0.90, 0.94}
\definecolor{maroon}{cmyk}{0,0.87,0.68,0.32}

\usepackage{pifont}% http://ctan.org/pkg/pifont

\renewcommand{\eqref}[1]{Eq.~(\ref{#1})}

\newcommand{\ellipses}{\textsc{Ellipses}}
\newcommand{\aapm}{\textsc{AAPM}}
\newcommand{\brats}{\textsc{BRATS}}
\newcommand{\fastmribrain}{\textsc{BRAIN}}
\newcommand{\fastmriknee}{\textsc{KNEE}}
\newcommand{\ddim}{{\rm DDIM}}

\newcommand{\tickYtopD}[1]{\raisebox{-1.5ex}[0ex][0ex]{#1}}

\newcommand{\tickPSNR}{\tickYtopD{PSNR}}

\newcommand{\tickRANK}[1]{\smash{RANK$=$}{$#1$}\hspace*{1em}}

\definecolor{C0}{rgb}{0.121569, 0.466667, 0.705882}
\definecolor{C1}{rgb}{1.000000, 0.498039, 0.054902}
\definecolor{C2}{rgb}{0.172549, 0.627451, 0.172549}
\definecolor{C3}{rgb}{0.839216, 0.152941, 0.156863}
\definecolor{C4}{rgb}{0.580392, 0.403922, 0.741176}
\definecolor{C5}{rgb}{0.549020, 0.337255, 0.294118}
\definecolor{C6}{rgb}{0.890196, 0.466667, 0.760784}
\definecolor{C7}{rgb}{0.498039, 0.498039, 0.498039}
\definecolor{C8}{rgb}{0.737255, 0.741176, 0.133333}
\definecolor{C9}{rgb}{0.090196, 0.745098, 0.811765}

\begin{document}

% ---------------------------------------------------------------
% TODO REVIEW: Replace with your title
\title{Deep Diffusion Image Prior for Efficient OOD Adaptation in 3D Inverse Problems}

% TODO REVIEW: If the paper title is too long for the running head, you can set
% an abbreviated paper title here. If not, comment out.
\titlerunning{DDIP: Diffusion OOD Adaptation in 3D Inverse Problems}

% TODO FINAL: Replace with your author list. 
% Include the authors' OCRID for the camera-ready version, if at all possible.
\author{Hyungjin Chung\orcidlink{0000-0003-3202-0893} \and
Jong Chul Ye\orcidlink{0000-0001-9763-9609}}

% TODO FINAL: Replace with an abbreviated list of authors.
\authorrunning{H.~Chung \& J.C.~Ye}
% First names are abbreviated in the running head.
% If there are more than two authors, 'et al.' is used.

% TODO FINAL: Replace with your institution list.
\institute{KAIST\\
\email{\{hj.chung,jong.ye\}@kaist.ac.kr}}

\maketitle

\begin{abstract}
Recent inverse problem solvers that leverage generative diffusion priors have garnered significant attention due to their exceptional quality. However, adaptation of the prior is necessary when there exists a discrepancy between the training and testing distributions. In this work, we propose deep diffusion image prior (DDIP), which generalizes the recent adaptation method of SCD~\cite{barbano2023steerable} by introducing a formal connection to the deep image prior. Under this framework, we propose an efficient adaptation method dubbed D3IP, specified for 3D measurements, which accelerates DDIP by orders of magnitude while achieving superior performance. D3IP enables seamless integration of 3D inverse solvers and thus leads to coherent 3D reconstruction. Moreover, we show that meta-learning techniques can also be applied to yield even better performance. We show that our method is capable of solving diverse 3D reconstructive tasks from the generative prior trained only with phantom images that are vastly different from the training set, opening up new opportunities of applying diffusion inverse solvers even when training with gold standard data is impossible. Code: \url{https://github.com/HJ-harry/DDIP3D}
\end{abstract}

\section{Introduction}
\label{sec:intro}

Diffusion models (DM) have shown remarkable performance as general inverse problem solvers~\cite{kadkhodaie2020solving,kawar2022denoising,chung2023diffusion}. The advantage naturally comes from the powerful capabilities of DMs to accurately model the prior distribution effectively learned from the collected training data. Unfortunately, there are numerous cases where the collection of high-quality gold standard data is not possible. For instance, it is difficult and expensive to collect large-scale data in medical imaging~\cite{zbontar2018fastmri,fabian2021data}, black hole imaging~\cite{event2019first} and cryo-EM imaging~\cite{zhong2021cryodrgn,gupta2021cryogan},  etc.
%it is impossible even to collect a {\em single} ground truth sample. 
Consequently, one either has to resort to phantoms~\cite{feng2023score,event2019first} for generative modeling, or leverage implicit priors~\cite{zhong2021cryodrgn}. In these challenging cases, one faces an out-of-distribution (OOD) problem arising from mismatched priors~\cite{renaud2023plug}, as the training data will be sufficiently different from the underlying true data distribution.

While it has been shown that DM-based inverse problem solvers (DIS) are less prone to distribution shifts~\cite{jalal2021robust,chung2022score}, the performance is significantly compromised~\cite{renaud2023plug,barbano2023steerable}, leading to large gaps in performance. In the reconstructed images, the discrepancy in the distributions is characterized as artifacts and hallucinations. As there exist provable bounds in the performance~\cite{renaud2023plug} when considering standard DIS with fixed parameters, the goal is to {\em adapt} the parameters of the diffusion model so that it better covers the true distribution, even when all we have access to is a degraded measurement.

\begin{wrapfigure}[18]{r}{0.5\textwidth}
\includegraphics[width=0.45\textwidth]{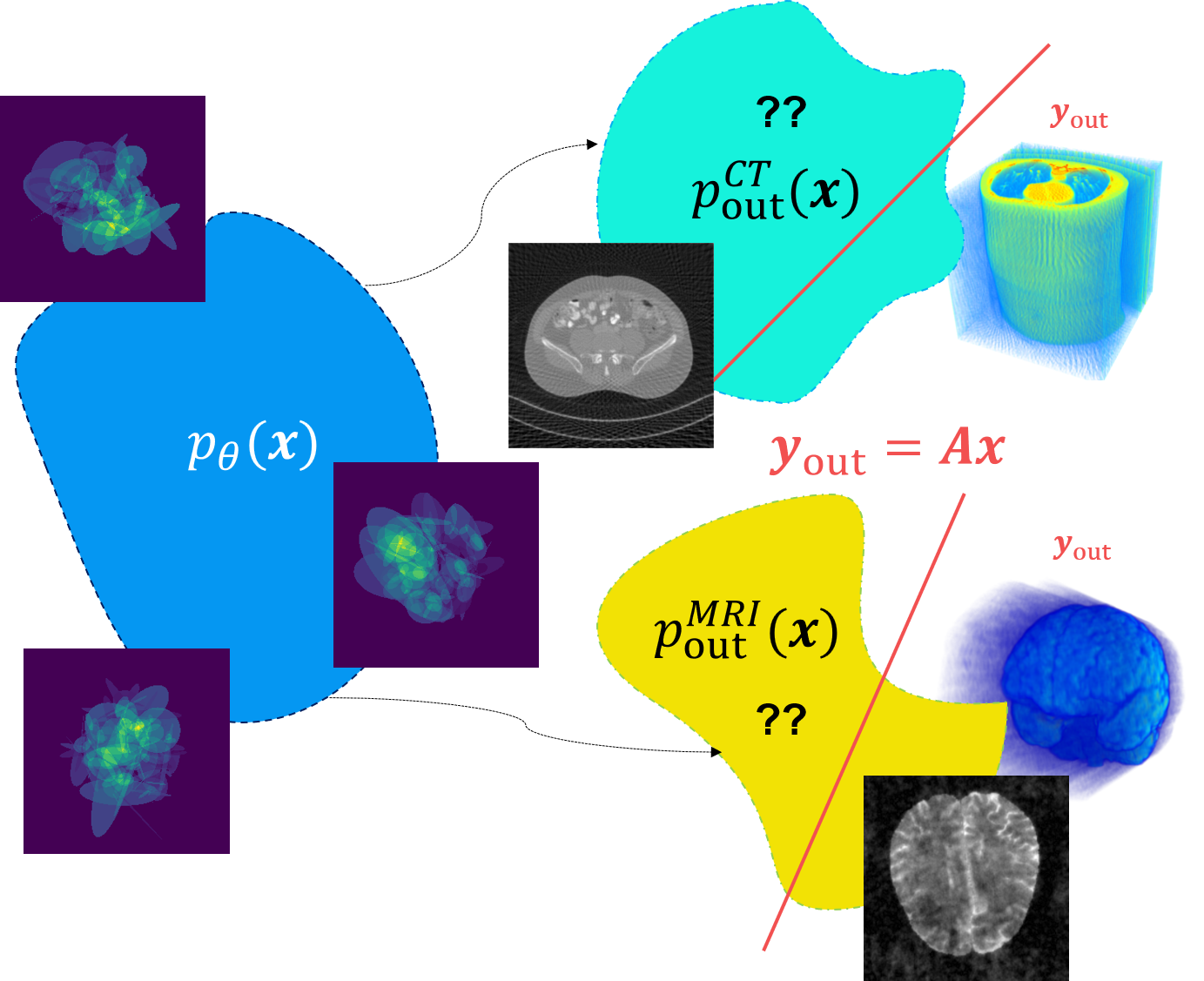}
\caption{OOD inverse problem setting. Pre-trained diffusion model learns $p_\theta(\x)$, but at test time we only have $\y_{\rm out}$ obtained from unknown OOD distributions, and aim to sample from $p_{\rm out}(\x|\y_{\rm out})$.}
\label{fig:problem_setting}
\end{wrapfigure}

Recently, SCD~\cite{barbano2023steerable} was proposed as the first work to demonstrate that one can adapt the parameters of the diffusion models during the reverse diffusion sampling process to steer the distribution towards the unknown. While it has been demonstrated that SCD is effective for boosting performance on several medical imaging inverse problems, there exist clear limitations of the work. First, the inverse problems considered such as computed tomography (CT) and magnetic resonance imaging (MRI) reconstruction problems are inherently volumetric. However, SCD requires different adapted parameters for every different slice of the 3D volume. As the adaptation framework slows down the inference time of DIS by $\times 5 \sim 30$ factor~\cite{barbano2023steerable}, the required reconstruction time for a $256^3$ volume requires more than 6 hours of inference time on a single commodity GPU. On top of computation and memory requirements growing with $\Oc(N)$ where $N$ is the number of 2D slices\footnote{This only considers the complexity along the slice dimension. The exact complexity will be presented in detail in Sec.~\ref{sec:exps}.}, it is also counter-intuitive that one needs a different set of parameters for adjacent slices, as the components are very similar. Ideally, the parameters should get better adapted as we increase the number of measurements, learning the common characteristics from the group. In contrast, we observe that the performance of SCD does not improve even if we try to gradually adapt the parameters with multiple samples.

In this work, we propose a novel framework that clarifies and solves all the aforementioned problems. Our contributions can be summarized as follows:

\begin{enumerate}
  \item We show a formal link between deep image prior (DIP)~\cite{ulyanov2018deep} and the adaptation framework introduced in SCD~\cite{barbano2023steerable}: when we correct for some factors in SCD, it is a multi-scale DIP constructed on the probability-flow ODE path~\cite{song2020score}. Under this view, SCD can be generalized as a deep diffusion image prior (\textbf{DDIP}).
  \item An efficient adaptation method specified for 3D inverse problem solving~\cite{chung2023diffusion,lee2023improving}, called \textbf{D3IP}, is proposed. It requires $\Oc(1)$ memory and compute, achieving significant acceleration while reaching performance on par with DDIP.
  \item We show that D3IP can also benefit from using a more advanced solver suited for 3D reconstruction by seamlessly combining it with 3D DIS (e.g. DiffusionMBIR~\cite{chung2023diffusion}). This lets us sample reconstructions that are spatially coherent across all the dimensions.
  \item We present a meta-learning~\cite{finn2017model,nichol2018reptile} algorithm for initializing a 3D adaptation model, and later fine-tuning to specific 2D slices, achieving higher performance by benefiting both from the volume statistics while also being able to fit a specific characteristic.
  \item Experimentally, we focus on the problem settings where we train a diffusion model on phantoms that can be generated on-the-fly, and hence can be considered fully unsupervised. This highlights the applicability to challenging imaging applications.
\end{enumerate}

\section{Background and Related Works}
\label{sec:background}

\subsection{Diffusion models for inverse problems}
\label{subsec:dis}

In inverse problems, we aim to retrieve the signal $\x$ from the measurement $\y$, where
\begin{align}
\label{eq:ip}
    \y = \Ab\x + \nb,\quad \y \in \Rd^m,\,\x\in\Rd^n,\,\Ab\in\Rd^{m \times n},\,\nb \sim \Nc(0,\sigma_y^2\Ib).
\end{align}
As the problem is ill-posed, a natural way to solve the problem is through Bayesian inference, aiming to sample from the posterior distribution $p(\x|\y) \propto p_{\rm data}(\x)p(\y|\x)$ by defining a suitable prior $p_{\rm data}(\x)$. A powerful way to do so is through generative modeling~\cite{goodfellow2014generative,kingma2013auto,ho2020denoising}, a predominant modern choice being diffusion models (DM)~\cite{sohl2015deep,ho2020denoising,song2020score}.

In DMs, a time variable $t \in [0, T]$ is introduced, where $p_0(\x_0) = p_{\rm data}(\x)$, and $p_T(\x_T)$ approaches an isotropic normal distribution $\Nc$ as $t \rightarrow T$ through a Gaussian noising process $p(\x_t|\x_0) = \Nc(\x_0, t^2\Ib)$. Interestingly, the reverse of such a process can be characterized as a stochastic differential equation (SDE) or a probability-flow ODE (PF-ODE)
\begin{align}
\label{eq:pf-ode}
    d\x_t = -t\nabla_{\x_t} \log p(\x_t)\,dt = \frac{\x_t - \Ed[\x_0|\x_t]}{t}\,dt, \quad p(\x_T) \sim \Nc,
\end{align}
where the second equality holds from Tweedie's formula $\Ed[\x_0|\x_t] = \x_t + t^2\nabla_{\x_t} \log p(\x_t)$~\cite{efron2011tweedie}. Due to this equivalence, one can train a denoiser $D_\theta$ through denoising score matching~\cite{vincent2011connection,song2019generative}, which estimates the denoised $\x_0$ from $\x_t$
\begin{align}
\label{eq:dsm}
    \theta^* = \argmin_\theta \Ed_{t \sim {\rm Unif}(\varepsilon, T), \x_t \sim p(\x_t|\x_0), \x_0 \sim p(\x_0)}\left[\|D_\theta(\x_t) - \x_0\|_2^2\right],
\end{align}
leading to $D_{\theta^*}(\x_t) \approx \Ed[\x_0|\x_t]$. Solving \eqref{eq:pf-ode} numerically by plugging $D_{\theta^*}$ in \eqref{eq:pf-ode} leads to sampling from the prior distribution $p(\x_0)$.

When aiming for {\em posterior} sampling, one can additionally condition the PF-ODE with $\y$, which reads
\begin{align}
\label{eq:posterior-pf-ode}
    d\x_t = -t\nabla_{\x_t} \log p(\x_t|\y)\,dt = \frac{\x_t - \Ed[\x_0|\x_t,\y]}{t}\,dt, \quad p(\x_T) \sim \Nc.
\end{align}
As $\Ed[\x_0|\x_t,\y]$ (equivalently $\nabla_{\x_t} \log p(\x_t|\y)$) is intractable, various methods have been proposed to approximate the posterior sampling process~\cite{chung2023diffusion,wang2023zeroshot,zhu2023denoising,chung2024decomposed}. DPS~\cite{chung2023diffusion} approximates
\begin{align}
\label{eq:dps}
    \Ed[\x_0|\x_t,\y] = \Ed[\x_0|\x_t] + t^2 \nabla_{\x_t} \log p(\y|\x_t)
    \approx \hat\x_{0|t} + \frac{t^2}{2\sigma_y^2} \nabla_{\x_t} \|\y - \Ab\hat\x_{0|t}\|_2^2,
\end{align}
with $\hat\x_{0|t} := D_{\theta^*}(\x_t)$\footnote{This notation is taken for simplicity. When emphasizing the dependence, we denote $\hat\x_{0|t}(\x_t; \theta^*)$}. As taking backprop w.r.t. $\x_t$ through the network is expensive and unstable~\cite{poole2023dreamfusion}, a way to avoid this was proposed in DDS~\cite{chung2024decomposed}, which uses
\begin{align}
\label{eq:dds}
    \Ed[\x_0|\x_t,\y] \approx \argmin_{\x_0} \frac{1}{2}\|\y - \Ab\x_0\|_2^2 + \frac{\gamma}{2}\|\x_0 - \hat\x_{0|t}\|_2^2,
\end{align}
solved with $M$-step conjugate gradient (CG), which we simply denote as $\code{CG}(\hat\x_{0|t}, M)$. Existing DIS can be considered as different ways of approximating the {\em empirical} conditional posterior, which we denote as $\Ed[\x_0|\x_t,\y] \approx D_\theta(\x_t|\y)$ (See Appendix~\ref{app:dis} for details and a review of existing methods).
Note that while we present an introduction to DMs in a variance exploding (VE) framework as in~\cite{karras2022elucidating}, extension to the variance preserving (VP) framework is straightforward~\cite{ho2020denoising}. In the following sections and all the experiments, we leverage the VP framework with standard notations adopted from \cite{ho2020denoising}. In most DIS, the predicted conditional posterior mean is used along with DDIM sampling~\cite{song2020denoising}
\begin{align}
\label{eq:ddim}
    \x_{t-1} = \ddim_\theta(\x_t,\eta) := \sqrt{\bar\alpha_{t-1}}D_\theta(\x_t|\y) + \sqrt{1 - \bar\alpha_{t-1}}\left(\eta\epsilonb + (1 - \eta)\epsilonb^\theta\right),
\end{align}
where $\eta \in [0, 1]$ controls the stochasticity and $\epsilonb^\theta$ is the predicted noise.

\subsection{Steerable Conditional Diffusion (SCD)}
\label{subsec:scd}

While solving \eqref{eq:posterior-pf-ode} with DIS approximations lets us perform posterior sampling, this is only true when the underlying prior distribution, and the distribution that $D_{\theta^*}$ models, are close enough (i.e. in-distribution). When $\y_{\rm out}$ is a measurement obtained from \eqref{eq:ip} with $\x_{\rm out} \sim p_{\rm out} \neq p_{\theta}$, it was shown that one can close the gap by running the following iteration
\begin{align}
\label{eq:scd}
    \text{for } t = T, \ldots, 1:
    \theta_{t-1} \gets &\argmin_{\theta_t}\|\y_{\rm out} - \Ab\code{CG}(\hat\x_{0|t}(\x_t; \theta_t),1)\|_2^2, \\
    \x_{t-1} \gets & \ddim_{\theta_{t-1}}(\code{CG}(\hat\x_{0|t}(\x_t; \theta_{t-1}),M), \eta)
\label{eq:scd_dds}
\end{align}
where additional LoRA~\cite{hu2021lora} parameters are introduced for adaptation to keep the original network parameters intact, and one-step CG was used to move the Tweedie denoised estimate closer towards $\y_{\rm out}$.
Adaptation of the network parameters in \eqref{eq:scd} is performed in between every DDS sampling step in \eqref{eq:scd_dds} for a fixed number of optimization steps $L$, which enables adaptation of the parameters on-the-fly during the sampling process. 

Nevertheless, one major limitation of SCD was the lack of understanding on why the algorithm works well in practice: Which design components are the key? What can we improve? Moreover, SCD requires running \eqref{eq:scd} for {\em every} different measurement $\y_{\rm out}^{i}$, where $i = 1, \dots, N$ denotes the slice index across a single volume with a total of $N$ slices. Re-initializing the parameters $\theta^{i}, \dots, \theta^{N}$, the process is $N$ times slower, memory expensive, and does not synergistically leverage information from adjacent slices.

\subsection{Related Works}
\label{subsec:related_works}

When the training data is unavailable, one of the most standard approaches in inverse imaging is the use of deep image prior (DIP)~\cite{ulyanov2018deep}
\begin{align}
\label{eq:dip}
    \theta^* = \argmin_{\theta} \|\y - \Ab G_\theta(\z)\|_2^2,~~ \z \sim \Nc(0, \Ib),
\end{align}
where $G_\theta$ produces the reconstruction, and some means of early stopping is used to prevent the network from too much overfitting. As neural networks favor output signals that lie in the natural data manifold~\cite{ulyanov2018deep}, and it can be considered a natural signal representation analogous to Fourier or Wavelets~\cite{hoyer2019neural}, optimizing for \eqref{eq:dip} leads to a decent reconstruction without any direct ground truth supervision.
Over the years, there have been numerous advances in each of these components: design of the loss function, early stopping criterion, network parametrization, and initialization~\cite{jo2021rethinking,heckel2018deep,baguer2020computed,liu2019image,barbano2022educated}. Despite the advances, DIP is still hard to optimize, and computationally demanding. For instance, applying DIP to a relatively small 3D data ($167^3$ resolution) requires about a day of training on a single RTX 3090 GPU~\cite{barbano2022educated}.

In this work, we extend the idea of DIP into the realm of diffusion inverse solvers, showing that we can yield a superior algorithm in speed, robustness, and quality by leveraging the merits of diffusion models. Two works that are perhaps the most related are Educated DIP~\cite{barbano2022educated} (EDIP) and Baguer \etal~\cite{baguer2020computed}. EDIP~\cite{barbano2022educated} shows that initializing $G_\theta$ with a network trained for reconstruction helps in convergence. \cite{baguer2020computed} proposes to use some initial reconstruction as an input to $G_\theta$, rather than some random vector $\z$. As will be later seen, our method naturally leverages both of these techniques by pivoting along the PF-ODE path.

On the other hand, there have been efforts to train a diffusion prior from corrupted measurements alone~\cite{kawar2023gsure,daras2023ambient}. However, this is only possible when the measurements satisfy strict conditions, and one needs abundant training data obtained from these identical conditions. Our work is orthogonal from such approaches as we are free from conditions, and our goal is not to train a new generative prior but to adapt an OOD prior for reconstruction.

\section{Proposed Method}
\label{sec:d3ip}

\begin{table}[!t]
\setlength{\tabcolsep}{2pt}
\newcolumntype{p}{>{\columncolor{BrickRed!10}}c}
\newcolumntype{s}{>{\columncolor{mylightblue}}c}
\newcolumntype{g}{>{\columncolor{gray!10}}c}
\centering
\resizebox{1.0\textwidth}{!}{% <------ Don't forget this %
\begin{tabular}{lgggscpcccpp}
\toprule
Solver & DDS & DDNM & DPS & \multicolumn{8}{c}{DDIP} \\
\cmidrule(lr){2-2}
\cmidrule(lr){3-3}
\cmidrule(lr){4-4}
\cmidrule(lr){5-11}
\cmidrule(lr){12-12}
{$\Ed[\x_0|\x_t,\y]$} & DDS & DDNM & DPS & \multicolumn{3}{c}{DDS} & \multicolumn{4}{c}{DDNM} & DPS\\
\cmidrule(lr){2-2}
\cmidrule(lr){3-3}
\cmidrule(lr){4-4}
\cmidrule(lr){5-7}
\cmidrule(lr){8-11}
\cmidrule(lr){12-12}
$K$ & - & - & - & 1 (SCD~\cite{barbano2023steerable}) & 3 & \textbf{5 (ours)} & 1 & 5 & 10 & \textbf{20 (ours)} & \textbf{1 (ours)}\\
\midrule
PSNR & 31.37 & 30.69 & 23.49 & 35.43 & 35.83 & \textbf{36.31} & 34.92 & 35.50 & 35.98 & \textbf{36.07} & \textbf{31.32}\\
SSIM & 0.887 & 0.870 & 0.608 & 0.911 & 0.921 & \textbf{0.922} & 0.902 & 0.911 & 0.920 & \textbf{0.920} & \textbf{0.841}\\
\bottomrule
\end{tabular}
}
\caption{
Sparse-view CT reconstruction results on $\ellipses \rightarrow \aapm$ OOD reconstruction by varying the solver, $\Ed[\x_0|\x_t,\y]$ approximation method, and the number of CG iterations $K$. \colorbox{gray!10}{Standard non-adapted DIS}, \colorbox{mylightblue}{SCD~\cite{barbano2023steerable}}, \colorbox{BrickRed!10}{Proposed DDIP method}.
}
\vspace{-0.5cm}
\label{tab:ddip_poc}
\end{table}

\subsection{Deep Diffusion Image Prior as a generalization of SCD}
\label{subsec:scd_dip}

Recall that DIP optimizes the network parameters with the fidelity loss in \eqref{eq:dip}.
As the optimization is held specific to $\y$, \eqref{eq:dip} aims to implicitly recover the posterior mean $\Ed[\x_0|\z,\y]$. On the other hand, as pointed out in Sec.~\ref{subsec:dis}, DIS at time $t$ during sampling, produces some estimate of the conditional posterior mean $\Ed[\x_0|\x_t,\y] \approx D_\theta(\x_t|\y)$. As $\x_t$ is equivalent to $\z$ at $t = T$ (initial noise), we can generalize DIP in \eqref{eq:dip} to multi-scale DIP over multiple noise scales
\begin{align}
\label{eq:ddip}
    \text{for } t = T, \ldots, 1:
    \theta_{t-1} \gets &\argmin_{\theta_t} \|\y - \Ab D_{\theta_t}(\x_t|\y)\|_2^2, \\
    \x_{t-1} \gets &\ddim_{\theta_{t-1}}(D_{\theta_{t-1}}(\x_t|\y), \eta)
\label{eq:ddip2}
\end{align}
where \eqref{eq:ddip2} denotes some diffusion inverse problem solver that proceeds by leveraging the adapted parameters $\theta_{t-1}$.
It is easy to see that \eqref{eq:ddip} is equivalent to \eqref{eq:dip} when $t = T$, but it is beneficial because 1) the optimization steps are performed in a coarse-to-fine manner starting from large noise scale, 2) the trajectory is pivoted along the original PF-ODE trajectory, providing a good initialization\footnote{As pointed out in Sec.~\ref{subsec:related_works}, our approach introduces a better initialization in the network parameters as in~\cite{barbano2022educated}, and also a better initialization in the input to the network~\cite{baguer2020computed} for $t < T$.}. We define a general method presented in \eqref{eq:ddip}, \eqref{eq:ddip2} as deep diffusion image prior (DDIP).

Revisiting \eqref{eq:scd} from this perspective immediately reveals that there exists a discrepancy between the conditional posterior mean estimation used for adaptation (with 1 CG iteration), and the conditional posterior mean estimation used for sampling (with $K > 1$ CG iterations). In Tab.~\ref{tab:ddip_poc}, we show that simply rectifying this error by using $K$ CG iterations also for adaptation significantly improves the performance, highlighting the advantage of our interpretation. Moreover, it now becomes clear that the parametrization of \eqref{eq:scd} is merely a design choice: any choice of producing the posterior mean can be used. 
To emphasize that DDIP is agnostic to the choice of the approximation, we show in Tab.~\ref{tab:ddip_poc} that DDIP consistently improves performance on OOD reconstruction regardless of the choices made.

\begin{figure}[t]
\begin{minipage}{.49\textwidth}
    \begin{algorithm}[H]
    \small
    \caption{DDIP}
    \label{alg:ddip}
    \begin{algorithmic}[1]
        \Require $\theta, D^t, N, T', \eta, \y$
        \State $\Lc(\x,\y,D_\theta) := \|\y - \Ab D_\theta(\x|\y)\|_2^2$
        \For{$i=1$ {\bfseries to} $N$}
             \State $\theta_{T'}^{(i)} \gets \theta$
             \State $\epsilonb \sim \Nc(0, \Ib)$
             \State $\x_T^{(i)} \gets \sqrt{\bar\alpha_{T'}}\Ab^\dagger \y + \sqrt{1 - \bar\alpha_{T'}}\epsilonb$
             \For{$t=T'$ {\bfseries to} $1$}
                 \State $\theta_{t-1}^{i} \gets \underset{\theta_t^i}{\arg\min}\Lc(\x_t^i,\y^i,D_{\theta_t^i}^t)$
                 \State $\hat\x_{0|t}^i \gets D_{\theta_{t-1}^i}^t(\x_t^i|\y^i)$
                 \State $\x_{t-1}^i \gets \ddim(\hat\x_{0|t}^i, \eta)$
             \EndFor
             \State {\bfseries return} $\x_0^{i}$
        \EndFor
    \end{algorithmic}
    \end{algorithm}
\end{minipage}
\begin{minipage}{.49\textwidth}
    \begin{algorithm}[H]
    \small
    \caption{D3IP}
    \label{alg:d3ip}
    \begin{algorithmic}[1]
        \Require $\theta, D^t, N, T', K, \eta, \Y$
        \State $\Lc(\x,\y,D_\theta) := \|\y - \Ab D_\theta(\x|\y)\|_2^2$
        \State $\theta_{T'} \gets \theta$
        \State $\epsilonb_1, \epsilonb_N \sim \Nc(0, \Ib)$
        \State $\epsilonb \gets {\rm slerp}(\epsilonb_1,\epsilonb_N,\frac{i}{N})$
        \State $\X_{T'} \gets \sqrt{\bar\alpha_{T'}}\Ab^\dagger\Y + \sqrt{1 - \bar\alpha_{T'}}\epsilonb$
        \For{$t=T'$ {\bfseries to} $1$}
            \State $\x_{t}^{\{i\}}, \y^{\{i\}} \sim {\rm MC}((\X_t,\Y), K)$
            \State $\theta_{t-1} \gets \underset{\theta_t}{\arg\min}\Lc(\x_t^{\{i\}},\y^{\{i\}},D_{\theta_t}^t)$
            \State $\hat\X_{0|t} \gets D_{\theta_{t-1}}^t(\X_t|\Y)$
            \State $\X_{t-1} \gets \ddim(\hat\X_{0|t}, \eta)$
        \EndFor
        \State {\bfseries return} $\X_0$
    \end{algorithmic}
    \end{algorithm}
\end{minipage}
\vspace{-0.5cm}
\end{figure}

\subsection{Extending DDIP to 3D}
\label{subsec:ddip_in_3d}

Let $\X \in \Rd^{n \times N} = [\x^{(1)}, \ldots, \x^{(N)}]$ a stacked 3D tensor with $N$ slices, and $\Y$ its corresponding measurement. In \cite{barbano2023steerable}, it was proposed to independently run \eqref{eq:scd} for every different slice, which requires a prohibitively long amount of time for adaptation/sampling. Concretely, with a $256^3$-sized volume that we consider in this work, it requires $90$ seconds per slice, which is $>6$ hours per volume.

\begin{figure}[t]
\centering
\includegraphics[width=\linewidth]{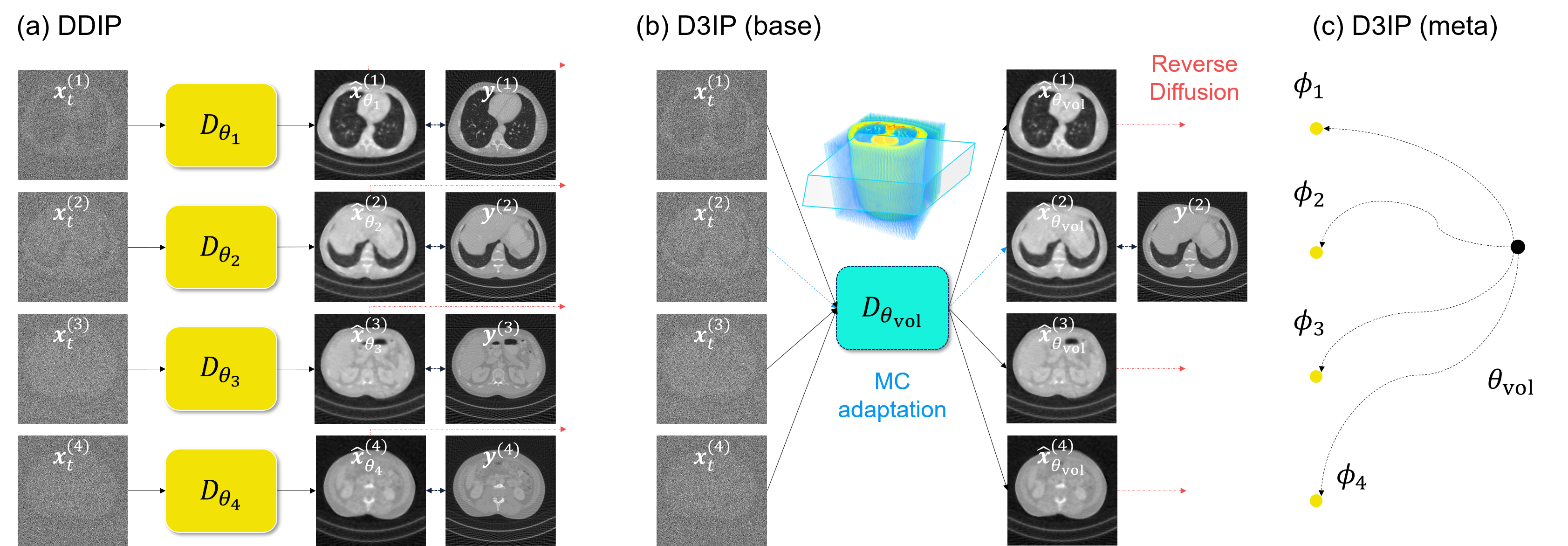}
\caption{OOD adaptation schemes in DIS. (a) DDIP/SCD performs {\em independent} adaptation across slices and requires $\Oc(N)$ compute \& memory. (b) D3IP (base) performs joint adaptation with stochastic gradients from MC sampling (blue dotted line) and requires $\Oc(1)$ compute \& memory. (c) $\theta_{\rm vol}$ adapted from D3IP (base) can be used as a meta-parameter to be further adapted to specific slices.}
\vspace{-0.5cm}
\label{fig:concept}
\end{figure}

\eqref{eq:ddip} is a sequential optimization problem that minimizes the fidelity loss in a multi-grid fashion for every slice $i$.
In SCD, the authors solve this optimization independently across all slices, as illustrated in Fig.~\ref{fig:concept} (a).
Instead, we propose to optimize \eqref{eq:ddip} so that the adaptation can be done synergistically in expectation (See Fig.~\ref{fig:concept} (b) for an illustration)
\begin{align}
\label{eq:d3ip}
    \text{for } t = T, \ldots, 1: \min_{\theta} \Ed_i\left[\|\y^{i} - \Ab D_\theta^t(\x_t^{i}|\y^{i})\|_2^2\right] \approx \frac{1}{K}\sum_{i=1}^K \|\y^{i} - \Ab D_\theta^{t}(\x_t^{i}|\y^{i})\|_2^2,
\end{align}
where $i$ denotes the index across the slices, and we can use $K$ Monte Carlo samples to compute the expectation. In practice, we find even using a small $K$ suffices for stable optimization and the performance plateaus for $K > 6$\footnote{See App.~\ref{app:ablation_studies} for further discussion.}, yielding a compute-effective algorithm.
The formulation of \eqref{eq:d3ip} lets us adapt the parameters that are suited for the whole 3D volume so that the memory and computation are reduced to $\Oc(1)$. 
We name our method D3IP (base), which will be shown to be improvable in specific settings, as we investigate in the following sections.
We highlight the difference of D3IP against DDIP in Alg.~\ref{alg:ddip},\ref{alg:d3ip}, where ${\rm MC}(\cdot, K)$ in L6 of Alg.~\ref{alg:d3ip} represents $K$-Monte Carlo sampling, and $\x^{\{i\}}$ denotes the sampled vectors stacked into a single tensor.
Surprisingly, not only is D3IP an order of magnitude cheaper and faster than DDIP, but it performs {\em better} than DDIP. This can be attributed to D3IP learning from a broader and richer context, whereas DDIP is only allowed to learn from a single slice of information.

\noindent
\textbf{Incorporating 3D DIS to D3IP.~}
Another big advantage of D3IP is that we can now seamlessly integrate 3D DIS methods~\cite{chung2023solving,lee2023improving} into our framework. As \cite{lee2023improving} would require adapting two diffusion models operating on different orientations, we choose DiffusionMBIR\cite{chung2023solving} accelerated with DDS\cite{chung2024decomposed} as our approximator (simply denoted DiffusionMBIR hereafter). Concretely, the approximation reads
\begin{align}
\label{eq:diffusionmbir_dds}
    D_\theta^{t}(\X_t|\Y) =  \argmin_{\X_0} \frac{1}{2}\|\y - \Ab\X_0\|_2^2 + \frac{\gamma}{2} \|\Tb\X_0\|_1,
\end{align}
where $\Tb$ computes the finite difference along the stacked slice dimension, and the optimization is solved with a few-step ADMM~\cite{boyd2011distributed} initialized with $\hat\X_{0|t}(\X_t; \theta_t)$ to induce proximal regularization. 

One caveat is that \eqref{eq:diffusionmbir_dds} requires computing total variation (TV) regularization for adjacent slices, which cannot be done if we randomly sample $K$ slices as done in D3IP (base). Instead, this can be effectively approximated by sampling $K < N$ neighboring slices, i.e. modifying L6 of Alg.~\ref{alg:d3ip} to
\begin{align}
\label{eq:d3ip_mbir_sample}
    \x_{t}^{i:i+K-1}, \y^{i:i+K-1} \sim {\rm MC}((\X_t,\Y), K).
\end{align}
We refer to this method with D3IP (mbir).
The regularization in \eqref{eq:diffusionmbir_dds} is already useful for the OOD setting as it is independent of training data, and 3D samplers typically provide a better estimate in 3D inverse problem settings. Consequently, we observe improved performance in the adapted setting by simply incorporating a 3D DIS.

Notice that for D3IP, we keep a single set of updated neural network parameters that are shared across the volume to save memory and compute. When one is willing to max out the performance by trading off more compute/memory, we show that one can do so by leveraging meta-learning. In fact, we reveal an interesting connection of D3IP to the widely used Reptile algorithm~\cite{nichol2018reptile} and show that one can easily leverage meta-learning into our framework. Details can be found in Appendix~\ref{subsec:meta_d3ip}.

\subsection{Technical Advances}
\label{subsec:technical_advances}

On top of the fundamental innovations, we propose several technical advances to the adaptation algorithm that yield faster and more robust optimization. These advances are orthogonal to the contributions that are proposed in the rest of the section, and can be applied to all adaptation methods.

\noindent
\textbf{Constraining the optimization horizon.~}
It is known that the empirical score functions exhibit problematic behavior in the end regimes~\cite{karras2022elucidating,kim2021soft} as the estimation tends to be inaccurate and volatile. Motivated from the score distillation sampling literature~\cite{poole2023dreamfusion,wang2023prolificdreamer}, we truncate the regime where we optimize for \eqref{eq:ddip} or \eqref{eq:d3ip}, so that for $t \notin [\zeta, T - \zeta]$, we only run standard DIS, with $\zeta = 40$ unless specified otherwise. We find that including adaptation outside this regime deteriorates the performance.

\noindent
\textbf{Initialization strategy.~}
It is standard practice to initialize DIS with random Gaussian noise~\cite{kawar2022denoising,chung2023diffusion,song2023pseudoinverseguided,wang2023zeroshot}. However, it is also known that due to the non-zero terminal SNR of diffusion models~\cite{lin2024common}, the low-frequency component of the initialization is carried out to the sample.
Due to this property, some works propose to initialize the Gaussian noise with low-frequency part replaced from a similar sample~\cite{wu2023freeinit}. In the case of inverse problems, this often comes for free. For instance, one can use the pseudo-inverse of the measurement.
Moreover, motivated by the initialization strategy for DIP~\cite{yoo2021time} and diffusion models~\cite{ge2023preserve} for video, which has correlations across time frames, we propose to sample two random noise vectors for the end slices and leverage the spherically interpolated (slerp) vector for the initialization noise. Summing up, our initialization reads
\begin{align}
\label{eq:y_init}
    \x_{T'}^{(i)} \gets \sqrt{\bar\alpha_{T'}}\Ab^\dagger\y^{(i)} + \sqrt{1- \bar\alpha_{T'}} {\rm slerp}(\epsilonb_1,\epsilonb_N,\frac{i}{N}), \quad \epsilonb_0,\epsilonb_N \sim \Nc(0, \Ib),
\end{align}
with $T' < T$ typically set to 980. This not only lets us start from a correlated initial noise vector with low-frequency components consistent with the measurement but also lets us avoid any instabilities arising from the non-zero terminal SNR problem~\cite{lin2024common}.

\section{Experiments}
\label{sec:exps}

We consider three canonical inverse problems in medical imaging: 3D sparse-view CT reconstruction (3D SV-CT), 3D MRI reconstruction (3D MRI), and multi-coil MRI reconstruction (CS-MRI), as these are the cases where the different measurement slices originate from the same volume. For the CT reconstruction task, we have a few hundred test slices originating from the same volume. For the first two tasks, we use a diffusion prior trained only on \ellipses~ phantoms~\cite{adler2018learned} generated on-the-fly, which are completely irrelevant to the target data distribution. We focus on such setting this leads to a completely unsupervised approach for a reconstructive task, requiring no collection of gold standard data.

For the MRI reconstruction task, we consider two different realistic cases: when the acquisition scheme is truly 3D and the volume consists of a few hundred slices, and when the acquisition scheme is 2D but the slices originate from the same volume, yielding a few ten slices. For all tasks, we consider variance preserving (VP) diffusion models trained under the ADM~\cite{dhariwal2021diffusion} framework unless specified otherwise.

\subsection{Experimental settings}
\label{subsec:exp_settings}

\noindent
\textbf{3D SV-CT.~}
Following \cite{barbano2023steerable}, we take a diffusion model trained on the \textsc{Ellipses} dataset~\cite{adler2018learned} and test it on a volume of the American Association of Physicists in Medicine (\textsc{AAPM}) grand challenge \cite{moen2021low} dataset, following the settings of~\cite{chung2022improving,chung2023solving}. This is a particularly useful and interesting setting, as the \textsc{Ellipses} dataset are phantoms that can be easily generated on-the-fly, requiring {\em no} collection of data, a realistic condition when acquiring high-quality ground truth images is impossible.
We consider parallel CT geometry with 60-angle measurements to be consistent with~\cite{barbano2023steerable}.

\begin{figure}[!t]
\centering
\includegraphics[width=1.0\linewidth]{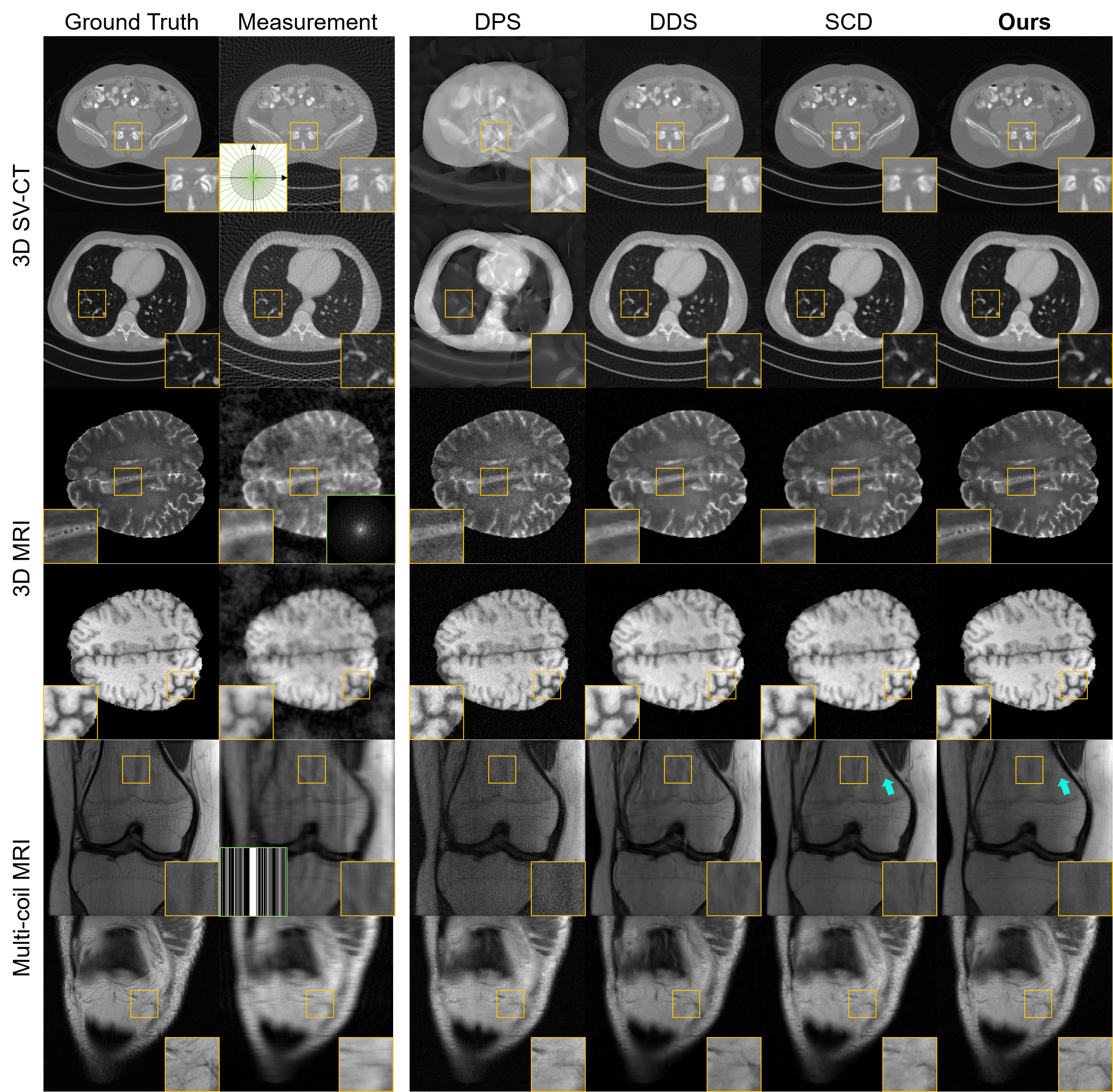}
\caption{Representative results on 3 different tasks. (row 1-2): 3D SV-CT, (row 3-4): 3D MRI, (row 5-6): CS-MRI. Comparison against DPS~\cite{chung2023diffusion}, DDS~\cite{chung2024decomposed}, and SCD~\cite{barbano2023steerable}. \textbf{Ours}: D3IP (base). Cyan arrows indicate regions of remaining artifacts even after adaptation with SCD. Green boxes illustrate the acquisition scheme of the measurement (acquisition angle, sub-sampling pattern).}
\vspace{-0.5cm}
\label{fig:main_results}
\end{figure}

\begin{table}[!t]
\centering
% \large
\setlength{\tabcolsep}{1pt}
\resizebox{1.0\textwidth}{!}{% <------ Don't forget this %
\begin{tabular}{lccc@{\hskip 15pt}ccc@{\hskip 15pt}ccc@{\hskip 15pt}cc}
\toprule
Method & \multicolumn{3}{c}{\thead{3D SV-CT \\ ($\ellipses \rightarrow \aapm$)}} & \multicolumn{3}{c}{\thead{3D MRI \\ ($\ellipses \rightarrow \brats$)}} & \multicolumn{3}{c}{\thead{Mult. coil CS-MRI \\ ($\fastmribrain \rightarrow \fastmriknee$)}} & Compute & Memory \\
\cmidrule{2-12}
& PSNR$\uparrow$ & SSIM$\uparrow$ & LPIPS$\downarrow$ & PSNR$\uparrow$ & SSIM$\uparrow$ & LPIPS$\downarrow$ & PSNR$\uparrow$ & SSIM$\uparrow$ & LPIPS$\downarrow$ & $^*K \ll T < N$\\
\cmidrule{2-4}
\cmidrule{5-7}
\cmidrule{8-10}
\cmidrule{11-12}
ADMM-TV & 25.35 & 0.783 & 0.220 & 23.89 & 0.508 & 0.441 & 24.64 & 0.689 & 0.340 & - & - \\
\midrule
\rowcolor{gray!10}
DDNM~\cite{wang2023zeroshot} & 23.55 & 0.765 & 0.241 & 14.59 & 0.281 & 0.753 & a19.35 & 0.354 & 0.493 & $\Oc(T)$ & $\Oc(1)$ \\
\rowcolor{gray!10}
DPS~\cite{chung2023diffusion} & 17.85 & 0.470 & 0.463 & 27.30 & 0.394 & 0.410 & 27.26 & 0.732 & 0.312 & $\Oc(T)$ & $\Oc(1)$ \\
\rowcolor{gray!10}
DDS~\cite{chung2024decomposed} & 27.65 & 0.805 & 0.188 & 24.59 & 0.532 & 0.339 & 28.36 & 0.741 & 0.278 & $\Oc(T)$ & $\Oc(1)$ \\
\rowcolor{gray!10}
DiffusionMBIR~\cite{chung2023solving} & 28.92 & 0.845 & 0.199 & 26.97 & 0.620 & 0.299 & - & - & - & $\Oc(T)$ & $\Oc(1)$ \\
\midrule
SCD~\cite{barbano2023steerable} & 32.91 & 0.904 & 0.184 & 26.00 & 0.561 & 0.338 & 29.01 & 0.752 & 0.269 & $\Oc(KNT)$ & $\Oc(N)$ \\
\rowcolor{BrickRed!10}
DDIP & 33.13 & 0.903 & 0.164 & 27.46 & 0.647 & 0.289 & \underline{29.11} & 0.775 & 0.246 & $\Oc(KNT)$ & $\Oc(N)$\\
\rowcolor{BrickRed!10}
D3IP~\textcolor{trolleygrey}{(base)} & 31.73 & 0.908 & 0.141 & 30.59 & 0.859 & 0.152 & 29.00 & \textbf{0.789} & \underline{0.233} & $\Oc(KT)$ & $\Oc(1)$ \\
\rowcolor{BrickRed!10}
D3IP~\textcolor{trolleygrey}{(mbir)} & \underline{33.69} & \textbf{0.919} & \textbf{0.133} & \textbf{33.89} & \textbf{0.907} & \textbf{0.103} & - & - & - & $\Oc(KT)$ & $\Oc(1)$ \\
\rowcolor{BrickRed!10}
D3IP~\textcolor{trolleygrey}{(meta)} & \textbf{33.96} & \underline{0.917} & \underline{0.136} & \underline{32.60} & \underline{0.877} & \underline{0.126} & \textbf{29.52} & \underline{0.779} & \textbf{0.216} & $\Oc(KNT)$ & $\Oc(N)$ \\
\bottomrule
\end{tabular}
}
\caption{
Quantitative measure of OOD Inverse problem solving on 3 main tasks. \colorbox{gray!10}{Non-adapted standard DIS}, the family of \colorbox{BrickRed!10}{proposed methods}.
}
\vspace{-0.5cm}
\label{tab:main_results}
\end{table}

\noindent
\textbf{3D MRI reconstruction}
We take the same \ellipses~ diffusion model used for 3D SV-CT, and adapt it to the multimodal brain tumor image segmentation benchmark (BRATS) 2018~\cite{menze2014multimodal}. The test set consists of 5 T1 contrast / 5 T2 contrast volumes. We consider variable density (VD) Poisson disc sampling pattern~\cite{dwork2021fast} ($\times 8$ acceleration) for 3D volume measurements in a single-coil setting.

\noindent
\textbf{2D Multi-coil MRI reconstruction.~}
A diffusion model trained on fastMRI~\cite{zbontar2018fastmri} \fastmribrain~ data was taken from~\cite{chung2024decomposed}\footnote{\url{https://github.com/HJ-harry/DDS}}. The evaluation was done on fastMRI \fastmriknee~ data, consisting of 10 test volumes and 294 slices. The measurement was simulated using the uniform 1D subsampling ($\times 4$ acceleration) with 8\% Auto Calibrating Signal (ACS) region, as proposed in~\cite{zbontar2018fastmri}. Since we consider multi-coil measurements in this task, the coil sensitivity maps are estimated using ESPiRiT~\cite{uecker2014espirit}.

For all inverse problems considered, we add $\sigma_y = 0.01$ measurement noise when simulating with forward operators. Further details can be found in App.~\ref{app:exp_details}.

\subsection{Diffusion samplers}
\label{subsec:diffusion_samplers}

For all DIS methods including the proposed method, we consider 50 NFE DDIM sampling with $\eta = 0.85$ unless specified otherwise. One exception is DPS~\cite{chung2023diffusion}, where we take 1000 NFE with $\eta = 1.0$ (i.e. DDPM sampling) to achieve satisfactory results.

\noindent
\textbf{Adaptation.~}
For the family of the proposed method (DDIP, D3IP), we take DDS~\cite{chung2024decomposed} as our estimator $D_\theta$ for $\Ed[\x_0|\x_t,\y]$, which runs 5 CG optimization step per diffusion denoising step. In 3D SV-CT and 3D MRI tasks, $K$ is set to 6, and for CS-MRI, $K = 3$. For D3IP (mbir), we use the approximation in \eqref{eq:diffusionmbir_dds}, which is solved with 5 ADMM iterations per diffusion step. D3IP (meta) runs D3IP in Alg.~\ref{alg:d3ip} with linearly decreasing step size from $\alpha = 1.0$ to $\alpha = 0.5$ in \eqref{eq:reptile}. The obtained meta-parameter $\theta_{\rm vol}$ is then fine-tuned with respect to each slice using DDIP in Alg.~\ref{alg:ddip} initialized from this meta-parameter.

\begin{wraptable}[7]{r}{0.4\textwidth}
\vspace{-0.7cm}
\centering
\setlength{\tabcolsep}{0.2em}
\resizebox{0.35\textwidth}{!}{% <------ Don't forget this %
\begin{tabular}{lll}
\toprule
{Method} & {PSNR $\uparrow$} & {SSIM $\uparrow$} \\
\cmidrule(lr){2-3}
DDIP(baseline) & 35.62 & 0.908\\
+ constrained horizon & 35.98 & 0.916\\
+ init. strategy & \textbf{36.31} & \textbf{0.922}\\
\bottomrule
\end{tabular}
}
\caption{
Improvements from configurations introduced in Sec.~\ref{subsec:technical_advances}.
}
\label{tab:technical_advances}
\end{wraptable}

\noindent
\textbf{Comparison methods.~}
We consider several strong DIS baselines: DDNM~\cite{wang2023zeroshot}, DPS~\cite{chung2023diffusion}, DDS~\cite{chung2024decomposed}, DiffusionMBIR~\cite{chung2023solving}, and SCD~\cite{barbano2023steerable} for comparison. In order to apply DDNM, one needs access to the pseudo-inverse operator, which is often hard to compute, or numerically unstable. We circumvent this by leveraging CG updates motivated from~\cite{chung2024decomposed}. We note that DiffusionMBIR is the only 3D-aware DIS, and SCD is the only adaptation-based DIS among the comparison methods. See App.~\ref{app:exp_details} for details.

\begin{figure}[!t]
\centering
\includegraphics[width=1.0\linewidth]{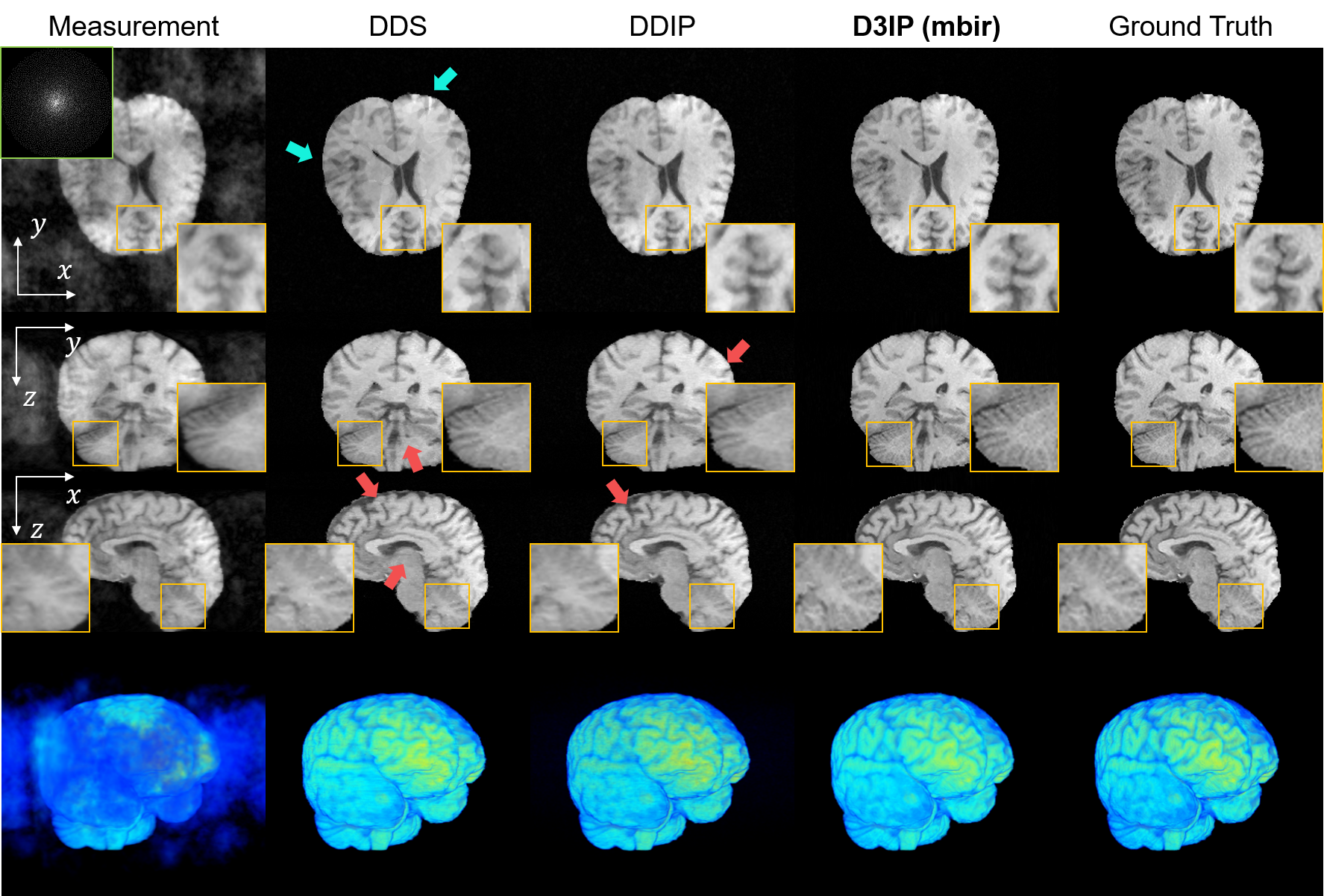}
\caption{3D-MRI reconstruction with DDS~\cite{chung2024decomposed}, DDIP, D3IP (mbir). Cyan and red arrows indicate artifacts from prior mismatch and slice-wise independent reconstruction, respectively. 1-4$^{\rm th}$ row: $xy, yz, xz$ slice, and 3D rendering.}
\vspace{-0.5cm}
\label{fig:main_results_3d}
\end{figure}

\subsection{Results}
\label{subsec:results}

\noindent
\textbf{Ablation study on the technical advances.~}
Before diving into the main results, we first conduct an ablation study introduced in Sec.~\ref{subsec:technical_advances}, as these techniques can be leveraged by all adaptation samplers. In Tab.~\ref{tab:technical_advances}, it is evident that the techniques provide improvements in the metrics, hence we keep the final configuration through all our experiments, including our proof-of-concept experiment presented in Tab.~\ref{tab:ddip_poc}.

\noindent
\textbf{Main Results.~}
In Tab.~\ref{tab:main_results}, we present a thorough comparison study on all three tasks that we consider in this work. Here, we see that D3IP not only dramatically reduces the computation cost of the adaptation, but also often results in a {\em better} reconstruction quality than DDIP. Specifically, this effect is most evidently seen in the 3D MRI task, where there is more than 3 dB difference in PSNR, and the perceptual LPIPS metric improves also by a large margin. Among all the tasks and the metrics considered, only the PSNR values of 3D SV-CT and Multi-coil CS-MRI fall short of DDIP. For the rest of the cases, D3IP outperforms DDIP, thanks to the richer information provided during the optimization process. In App.~\ref{app:ablation_studies}, we further show the intriguing properties of D3IP: robustness to the choice of hyperparameters, and higher capacity.

In Fig.~\ref{fig:main_results}, we compare the qualitative results of our proposed D3IP (base) solver against SOTA DIS - DPS, DDS, and the previous adapted sampler SCD. It is clear that the results obtained with DPS and DDS are contaminated with artifacts and hallucinations that originate from the training data (elliptical phantoms). While SCD greatly alleviates these artifacts, the details are typically blurred out, erasing structures that can even be seen from the measurement (e.g. 1st row of Fig.~\ref{fig:main_results}). Notably, this happens even without any TV regularization that was used in~\cite{barbano2023steerable}. Introducing such regularization yields even blurrier results. In contrast, the proposed method is capable of restoring crisp features without such blurring. Moreover, D3IP further eliminates leftover artifacts that are still apparent in SCD, as pointed out with cyan arrows.

\begin{wrapfigure}[15]{r}{0.6\textwidth}
\vspace{-0.5cm}
\includegraphics[width=0.55\textwidth]{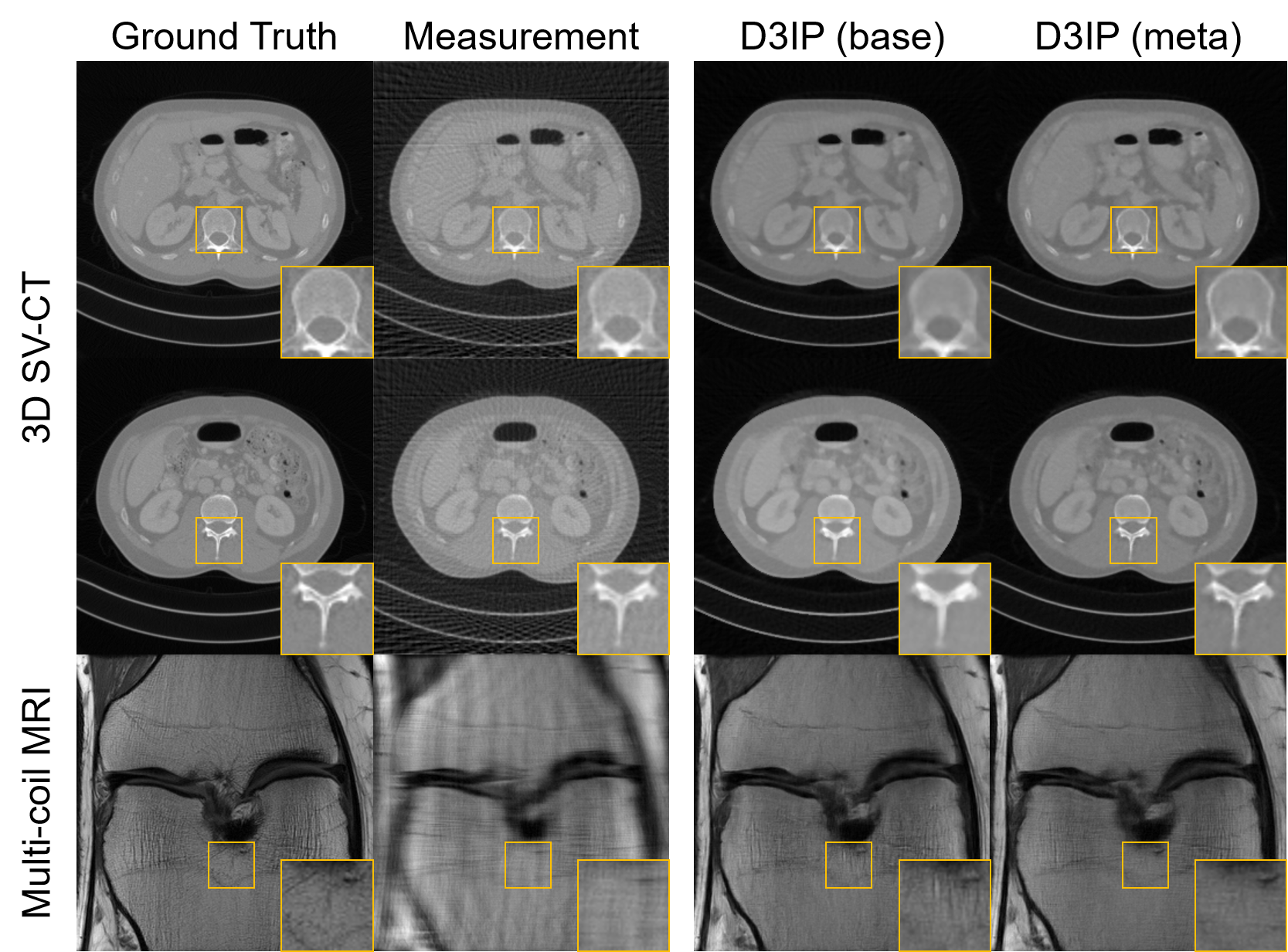}
\caption{Comparison of reconstructions with D3IP (base) and D3IP (meta).}
\label{fig:main_results_meta}
\end{wrapfigure}

It should be noted that all these advances take place while also being much faster and cheaper in memory. On a single 3090 RTX GPU, for a 256$^3$ volume, D3IP (base) requires 40 minutes, whereas SCD requires $\sim$6.2 hours, as it should be conducted slice-wise. Moreover, by using D3IP, we can reduce the memory requirements of the LoRA parameters from 2.9 Gb to 14.5 Mb as we only need to keep a single set of parameters for the whole volume, rather than specific parameters for each slice. 

\noindent
\textbf{Incorporating DiffusionMBIR.~}
While D3IP (base) enables the use of average gradients computed from multiple slices, it does not guarantee smooth reconstruction across slices, which can be effectively counteracted by leveraging 3D solvers. In Fig.~\ref{fig:main_results_3d}, we visualize the 3D rendering as well as the coronal and the sagittal slices of the reconstruction, highlighting the advantage of the proposed method by allowing a plug-and-play incorporation of 3D DIS.

\noindent
\textbf{Meta-learning further improves D3IP.~}
By initializing a meta-parameter $\theta_{\rm vol}$ and further adapting the parameters toward specific slices, we see in Fig.~\ref{fig:main_results_meta} and Tab.~\ref{tab:main_results} that we can achieve even better performance by trading off compute. In contrast, gradually adapting the parameters across different slices with SCD does not lead to noticeable improvements, compared to the case where the weights are re-initialized for every slice.

\section{Conclusion}
\label{sec:conclusion}

In this work, we proposed DDIP, a method of adapting diffusion priors trained on OOD distribution for solving inverse imaging tasks. We clarified that DDIP is a generalization of DIP constructed on the PF-ODE trajectory of diffusion, which makes the method much stabler than standard DIP. Focusing on 3D inverse problems, we proposed D3IP, which significantly reduces the computation cost while achieving better performance. We showed that D3IP can be further enhanced by incorporating 3D inverse problem solvers, or by leveraging meta-learning to induce a meta-parameter, then fine-tuning to specific 2D measurements. Notably, our work improves previous approaches in reconstruction quality, closing the gap against in-distribution DIS; it accelerates the speed of the algorithm to a regime where it will be practical for real-world usage; it enhances the interpretability of the algorithm and clarifies why the method works. We believe that our work can become the ground for many inverse imaging applications where unsupervised reconstruction remains crucial, such as in biomedical imaging and astronomy.

\section*{Acknowledgements}

This work was supported by the National Research Foundation of Korea under Grant RS-2024-00336454, RS-2023-00262527, by Field-oriented Technology Development Project for Customs Administration funded by the Korea government (the Ministry of Science \& ICT and the Korea Customs Service) through the National Research Foundation (NRF) of Korea under Grant NRF2021M3I1A1097910, by Culture, Sports and Tourism R\&D Program through the Korea Creative Content Agency grant funded by the Ministry of Culture, Sports and Tourism in 2023, and by the Institute of Information \& communications Technology Planning \& Evaluation (IITP) grant funded by the Korea government(MSIT)  
(No.2019-0-00075, Artificial Intelligence Graduate School Program(KAIST))

% ---- Bibliography ----
%
% BibTeX users should specify bibliography style 'splncs04'.
% References will then be sorted and formatted in the correct style.
%
\bibliographystyle{splncs04}
\bibliography{refs}

\clearpage 
\appendix

\section{Diffusion model-based inverse problem solvers}
\label{app:dis}

In this section, we give an overview of the different diffusion model-based inverse problem solvers (DIS), and what different approximations are used for each method, focusing on the methods that are used as comparison methods and how they are practically implemented. Recall the PF-ODE for posterior sampling in the VE setup (\eqref{eq:posterior-pf-ode})
\begin{align}
    d\x_t = -t\nabla_{\x_t} \log p(\x_t|\y)\,dt = \frac{\x_t - \Ed[\x_0|\x_t,\y]}{t}\,dt, \quad p(\x_T) \sim \Nc,
\end{align}
where we clearly see that the PF-ODE is governed by the conditional posterior mean $\Ed[\x_0|\x_t,\y]$. The Tweedie's formula~\cite{efron2011tweedie} states that
\begin{align}
\label{eq:tweedie_base}
    \Ed[\x_0|\x_t] = \x_t + t^2\nabla_{\x_t} \log p(\x_t) \approx \x_t + t^2 \s_{\theta^*}(\x_t),
\end{align}
where $\s_{\theta^*}(\cdot)$ is a diffusion model parametrized as a score function and trained through denoising score matching (DSM)~\cite{hyvarinen2005estimation}. Alternatively, one can directly utilize the parametrization in \eqref{eq:tweedie_base} and treat the neural network as a denoiser $\Ed[\x_0|\x_t] \approx D_{\theta^*}(\x_t)$. Leveraging Bayes rule
\begin{align}
\label{eq:bayes}
    \nabla_{\x_t} \log p(\x_t|\y) = \nabla_{\x_t} \log p(\x_t) + \nabla_{\x_t} \log p(\y|\x_t),
\end{align}
we have
\begin{align}
    \Ed[\x_0|\x_t,\y] &= \x_t + t^2\nabla_{\x_t} \log p(\x_t|\y) \\
    &= \x_t + t^2 \nabla_{\x_t} \log p(\x_t) + t^2 \nabla_{\x_t} \log p(\y|\x_t) \\
    &= \Ed[\x_0|\x_t] + t^2 \nabla_{\x_t} \log p(\y|\x_t) \\
    &\approx D_{\theta^*}(\x_t) + t^2 \nabla_{\x_t} \log p(\y|\x_t).
\end{align}
Here, we see that approximating $\Ed[\x_0|\x_t,\y]$ is equivalent to approximating $\nabla_{\x_t} \log p(\y|\x_t)$. Consequently, the methods that aim to modify the Tweedie denoised estimate~\cite{wang2023zeroshot,chung2024decomposed,zhu2023denoising} and the methods that aim to approximate $\nabla_{\x_t} \log p(\y|\x_t)$~\cite{jalal2021robust,chung2023diffusion} can be understood in a unified framework. In the following, we review representative DIS methods that are also used as comparison methods for the study. For each method, we also discuss their implementation details and the hyper-parameters used.

\noindent
\textbf{DDNM~\cite{wang2023zeroshot}~}
aims for projection to match perfect measurement consistency via range-null space decomposition
\begin{align}
\label{eq:ddnm_approx}
    \Ed[\x_0|\x_t,\y] \stackrel{\text{(DDNM)}}{\approx} (\Ib - \Ab^\dagger \Ab)D_{\theta^*}(\x_t) + \Ab^\dagger \yb
\end{align}
As for many medical imaging applications, directly applying the pseudo-inverse operator is either infeasible or unstable, we can instead solve $\Ab^\top\Ab D_{\theta^*}(\x_t) = \yb$ through conjugate gradient (CG), i.e. $\code{CG}(D_{\theta^*}(\x_t), M)$ with a sufficiently large $M$. For our experiments, we set $M = 30$.

\noindent
\textbf{DPS~\cite{chung2023diffusion}~}
approximates
\begin{align}
\label{eq:dps_approx_verbose}
    \nabla_{\x_t} \log p(\y|\x_t) \stackrel{\text{(DPS)}}{\approx} \rho \nabla_{\x_t} \|\y - \Ab D_{\theta^*}(\x_t)\|_2,
\end{align}
where the Jensen gap is induced by using the empirical posterior mean $D_{\theta^*}(\x_t)$ in the place of $\x_t$. Empirically, it was found that using a static step size with a non-squared norm as in \eqref{eq:dps_approx_verbose} was shown to be effective. In our experiments, we use $\rho = 0.5$. Note that this is equivalent to the following approximation
\begin{align}
\label{eq:dps_approx_verbose2}
    \Ed[\x_0|\x_t,\y] \stackrel{\text{(DPS)}}{\approx} \Ed[\x_0|\x_t] + t^2\rho \nabla_{\x_t} \|\y - \Ab D_{\theta^*}(\x_t)\|_2.
\end{align}

\noindent
\textbf{DDS~\cite{chung2024decomposed}~}
performs the following proximal optimization
\begin{align}
    \Ed[\x_0|\x_t,\y] \stackrel{\text{(DDS)}}{\approx} \argmin_{\x_0}
    \frac{\gamma}{2}\|\y - \Ab\x_0\|_2^2 + \frac{1}{2}\|\x_0 - D_{\theta^*}(\x_t)\|_2^2,
\end{align}
which amounts to solving the following linear system of equation
\begin{align}
\label{eq:dds_prox}
    (\gamma \Ab^\top\Ab + \Ib)\x_0^* = \gamma\y + D_{\theta^*}(\x_t)
\end{align}
through CG. To prevent falling off from the data manifold, \eqref{eq:dds_prox} is solved in a coarse manner by using a small number of iterations. We use $M = 5, \gamma=5$ for all our experiments, as well as the adaptation solvers that are based on the DDS approximation.

\noindent
\textbf{DiffusionMBIR~\cite{chung2023solving}~}
is a method that explicitly induces smooth transition across otherwise-independent slices by using the following optimization
\begin{align}
    \Ed[\X_0|\X_t,\Y] \stackrel{\text{(DiffusionMBIR)}}{\approx} \argmin_{\X_0}
    \frac{1}{2}\|\y - \Ab\x_0\|_2^2 + \frac{\lambda}{2}\|\Tb D_{\theta^*}(\x_t)\|_1,
\end{align}
\begin{wraptable}[12]{r}{0.4\textwidth}
\vspace{-0.5cm}
\centering
\setlength{\tabcolsep}{0.2em}
\resizebox{0.35\textwidth}{!}{% <------ Don't forget this %
\begin{tabular}{lcc@{\hskip 15pt}cc}
\toprule
{} & \multicolumn{2}{c}{\textbf{3D SV-CT}} & \multicolumn{2}{c}{\textbf{3D MRI}} \\
\cmidrule(lr){2-3}
\cmidrule(lr){4-5}
$K$ & PSNR$\uparrow$ & SSIM$\uparrow$ & PSNR$\uparrow$ & SSIM$\uparrow$ \\
\midrule
1 & 34.82 & 0.915 & 31.48 & 0.852 \\
2 & 35.05 & 0.922 & 31.49 & \textbf{0.853} \\
3 & 35.09 & 0.920 & 31.50 & 0.852 \\
4 & 35.16 & 0.923 & 31.53 & 0.852 \\
6 & \textbf{35.60} & \textbf{0.926} & \textbf{31.53} & 0.852 \\
12 & 35.46 & 0.925 & 31.51 & 0.850 \\
24 & 35.56 & 0.924 & 31.52 & 0.849 \\
\bottomrule
\end{tabular}
}
\caption{
Effect of varying $K$ in D3IP.
}
\label{tab:mc_ablation}
\end{wraptable}
which can be solved with ADMM~\cite{chung2023solving,boyd2011distributed}. To solve the method with ADMM, we have $\rho$, a constant set for the method of multipliers, and $\gamma$, a constant that weights the importance of the TV regularization. For both 3D CT and 3D SV-CT, we run grid search over these two parameters and use the best combination for both DiffusionMBIR and D3IP (mbir). For 3D MRI, we use $\rho = 1e-3, \lambda = 1e-5$. For 3D SV-CT, we use $\rho=0.5, \lambda=0.01$. For every timestep $t$, we run 5 ADMM steps and 5 inner-CG steps for optimization.

\section{Meta-learning D3IP}
\label{subsec:meta_d3ip}

The main motivation of D3IP was to reduce the memory and computation cost for OOD adaptation of diffusion models. 
Nevertheless, we see in practice that D3IP does not always achieve better performance than SCD, which can be attributed to the fact that the parameters are being adapted to an {\em average} direction. Negative transfer classically arising in multi-task optimization~\cite{crawshaw2020multi,desideri2012multiple} can lead to a suboptimal result for each slice.

On the other hand, consider meta-learning~\cite{nichol2018reptile,finn2017model}, where the objective is to find a good {\em meta} parameter that can be quickly adapted to different tasks that one considers. 
Interestingly, our formulation can be cast as a meta-learning problem when we view the optimization problem with respect to each slice, a single task. Recall that the Reptile algorithm~\cite{nichol2018reptile}, at the $t$-th step going on to $t-1$-th step, follows
\begin{align}
\label{eq:reptile}
    \tilde\theta_t \gets U_{\{i\}}^c(\theta_t), \quad \theta_{t-1} \gets \theta_t + \alpha(\tilde\theta_t - \theta_t), \quad \alpha_t \in (0, 1],
\end{align}
where $U_{\{i\}}^c(\cdot)$ is $c$-step gradient update using an optimizer where the sampled tasks are in $\{i\}$. Interestingly, under this view, Alg.~\ref{alg:d3ip} is already the Reptile algorithm~\cite{nichol2018reptile} where the optimization problem of L7 is solved with $U_{\{i\}}^c(\cdot)$,  and a constant step size $\alpha_t = 1.0$ is chosen, corresponding to making full updates. Following~\cite{nichol2018reptile}, we define D3IP (meta) by choosing a value of $\alpha_t$ to linearly decay starting from $1.0$ at the initial iteration, providing a better initialization point to be further fitted to each slice. We refer to the meta-parameter as $\theta_{\rm vol}$.

After running the meta-learning algorithm, when one is willing to trade-off more compute with better performance, we can further fine-tune our adapted meta-parameter $\theta$ on respective slices to obtain a parameter set $\{\phi_1, \ldots, \phi_N\}$ with the usual DDIP, initializing Alg.~\ref{alg:ddip} for every slice from $\theta_{\rm vol}$, achieving higher quality reconstruction than standard DDIP without meta-learning. The illustration of the idea is shown in Fig.~\ref{fig:concept} (c).

\section{Experimental details}
\label{app:exp_details}

\begin{table}[!t]
\centering
\setlength{\tabcolsep}{1pt}
\resizebox{1.0\textwidth}{!}{% <------ Don't forget this %
\begin{tabular}{l@{\hskip 15pt}ccccc@{\hskip 15pt}ccccc@{\hskip 15pt}cccc}
\toprule
& \multicolumn{5}{c}{\textbf{3D SV-CT}} & \multicolumn{5}{c}{\textbf{3D MRI}} & \multicolumn{4}{c}{\textbf{CS-MRI}} \\
\cmidrule(lr){2-6}
\cmidrule(lr){7-11}
\cmidrule(lr){12-15}
\textbf{Method} & SCD~\cite{barbano2023steerable} & DDIP & \thead{D3IP \\ (base)} & \thead{D3IP \\ (mbir)} & \thead{D3IP \\ (meta)} & SCD~\cite{barbano2023steerable} & DDIP & \thead{D3IP \\ (base)} & \thead{D3IP \\ (mbir)} & \thead{D3IP \\ (meta)} & SCD~\cite{barbano2023steerable} & DDIP & \thead{D3IP \\ (base)} & \thead{D3IP \\ (meta)} \\
\midrule
$K$ & 1 & 1 & 6 & 6 & 6 & 1 & 1 & 6 & 6 & 6 & 1 & 1 & 3 & 3 \\
$L$ & 10 & 10 & 10 & 10 & 10 & 10 & 10 & 10 & 5 & 10 & 5 & 5 & 5 & 5 \\
lr & 1e-3 & 1e-3 & 1e-3 & 1e-3 & 1e-3 & 1e-5 & 1e-5 & 1e-4 & 1e-5 & 1e-5 & 1e-3 & 1e-3 & 1e-3 & 1e-3 \\
\bottomrule
\end{tabular}
}
\caption{
Hyperparameters for the adaptation DIS methods presented in this study.
}
\label{tab:hparam}
\end{table}

Following SCD~\cite{barbano2023steerable}, we expand the parameters of the original pre-trained diffusion model with LoRA parameters injected to all the convolutional residual blocks and the attention blocks with rank 4, unless specified otherwise. The newly introduced parameters consist of approximately 0.61\% of the total number of parameters. When running the adaptation, we use the AdamW optimizer~\cite{loshchilov2017decoupled} with default parameters. The three parameters specified for the adaptation, $K$ (the number of MC samples used for stochastic gradients), $L$ (the number of iterations for optimizing \eqref{eq:ddip} or \eqref{eq:d3ip}), and the learning rate, are presented in Tab.~\ref{tab:hparam}.

\section{Ablation studies}
\label{app:ablation_studies}

\noindent
\textbf{Monte Carlo sampling for D3IP.~}
D3IP in Alg.~\ref{alg:d3ip} utilizes $K$-MC sampling, where the gradients would be more accurately computed when we set $K$ to the number of slices that we consider in the volume (e.g. 256 for 3D SV-CT). Nevertheless, as we are mostly concerned with compute-efficient scenario where we utilize a single GPU, setting $K > 3$ yields OOM issues. In Tab.~\ref{tab:mc_ablation}, we inspect whether increasing the value of $K$ indeed leads to better reconstruction performance. 

\begin{figure}[!t]
    \centering
    \begin{subfigure}[t]{0.5\textwidth}
    \adjustbox{width=\columnwidth,valign=c}{
        \begin{tikzpicture}
    	\begin{axis}[
    	height=5cm, width=6cm,font=\tiny,
    	xmin={4}, xmax={40}, xtick={4, 8, 12, 16, 20, 24, 28, 32, 36, 40}, xticklabels={\tickRANK{4}, $8$, $12$, $16$, $20$, $24$, $28$, $32$, $36$, $40$},
    	ymin={34.0}, ymax={36.0}, ytick={34.1, 34.5, 34.9, 35.3, 35.7, 36.0},          yticklabels={$34.1$, $34.5$, $34.9$, $35.3$, $35.7$, \tickPSNR},
            grid={major}, legend style={font=\tiny, at={(0.75, 0.3)}},
    	]
    	\addplot[C0,mark=o,mark size=1.5pt] coordinates {
    		(4, 34.17)
    		(8, 34.56)
    		(12, 34.92)
    		(16, 35.20)
    		(20, 35.63)
    		(24, 35.52)
                (28, 35.57)
                (32, 35.20)
                (36, 35.47)
                (40, 35.10)
    	};
    	\addlegendentry{DDIP}
    	\addplot[C3,mark=diamond,mark size=1.5pt] coordinates {
    		(4, 34.23)
    		(8, 34.49)
    		(12, 34.90)
    		(16, 35.08)
    		(20, 35.24)
    		(24, 35.25)
                (28, 35.39)
                (32, 35.96)
                (36, 35.61)
                (40, 35.57)
    	};
    	\addlegendentry{D3IP}
    	\end{axis}
        \end{tikzpicture}
        }
    \caption{PSNR values in 3D SV-CT taks by varying the LoRA~\cite{hu2021lora} rank}
    \label{fig:lora_rank}
    \end{subfigure}
%%%%%%%%%%%%%%%%%%%%%%%%%%%%%%%%%%%%%%%%%%%%%%%%%%%%%%%%%
    \begin{subtable}[t]{0.4\textwidth}
        \setlength{\tabcolsep}{0.2em}
        \resizebox{1.0\textwidth}{!}{% <------ Don't forget this %
        \begin{tabular}{lccc@{\hskip 15pt}ccc}
        \toprule
        {} & \multicolumn{3}{c}{\textbf{DDIP}} & \multicolumn{3}{c}{\textbf{D3IP}} \\
        \cmidrule(lr){2-4}
        \cmidrule(lr){5-7}
        \backslashbox{\textbf{lr}}{\textbf{$L$}} & 3 & 5 & 10 & 3 & 5 & 10\\
        \midrule
        1e-3 & 29.95 & 30.24 & 27.63 & 30.76 & 30.75 & 30.76 \\
        1e-4 & 30.46 & 30.40 & 30.21 & 30.41 & 30.63 & 30.66 \\
        \bottomrule
        \end{tabular}
        }
        \caption{
        PSNR values in 3D MRI task by varying the hyperparameters for adaptation.
        }
        \label{tab:robustness_variation}
     \end{subtable}
\caption{Ablation studies performed with 3D SV-CT and 3D MRI tasks.}
\label{fig:appendix_ablation}
\end{figure}

\noindent
\textbf{Robustness to variation in adaptation.~}
In standard DIP~\cite{ulyanov2018deep}, early stopping as well as finding the right learning rate is of crucial importance. Running the optimization for too long or using too large of a learning rate could easily lead to divergence. While we find that a similar trend persists for the proposed method, in Tab.~\ref{tab:robustness_variation} we show that D3IP is more robust to the variations in the hyper-parameters: $L$ (number of iterations per timestep $t$) and learning rate. For instance, with lr$=$1e-3 and $L=10$, DDIP fails to provide a better reconstruction than DDS, whereas D3IP is stable.
This is another advantage of D3IP that would be of great importance for practical usage.

\noindent
\textbf{LoRA adaptation.~}
While our default LoRA rank was set to 4 for all the experiments that were conducted, we can increase the LoRA rank if we wish the network to have a higher capacity to be fitted better to the given measurement. In Fig.~\ref{fig:appendix_ablation} (a), we perform an ablation study on 3D SV-CT on DDIP and D3IP to inspect the optimal LoRA rank for each. Aligned with the intuition, we can see that the optimal rank of D3IP is larger than that of DDIP, as it has more information to learn from. The PSNR values plateau faster than D3IP. As an additional note, in our initial experiments, we found that full model finetuning achieves worse performance than LoRA adaptation and that introducing LoRA parameters not only to the attention blocks but to the residual blocks leads to better performance.

\section{Discussion and Limitations}
\label{sec:discussion}

The proposed method adapts the parameters of the diffusion models on-the-fly, consequently using different parameters $\theta_t$ at every timestep. Hence, in order to replicate the behavior {\em after} sampling, one needs to store different $\theta_t$ for all $t$. When we simply use the final LoRA parameters $\theta_0^*$ and run standard DIS, the performance degrades. As we impose no regularization to the learned LoRA parameters during adaptation, the in-distribution reconstruction performance typically degrades due to catastrophic forgetting~\cite{mccloskey1989catastrophic,kirkpatrick2017overcoming}. This can be only counteracted by manually turning the LoRA parameters on and off depending on the membership of the measurement, which is cumbersome. These issues stem from the fact that the adaptation is done online during reverse diffusion sampling, as opposed to offline adaptation methods often leveraged in diffusion model literature, where parameter expansion is used~\cite{ruiz2023dreambooth,zhang2023adding}.
We leave this venue for a future direction of research.

\section{Further experimental results}
\label{app:further_exp_results}

\begin{figure}[!t]
\centering
\includegraphics[width=\linewidth]{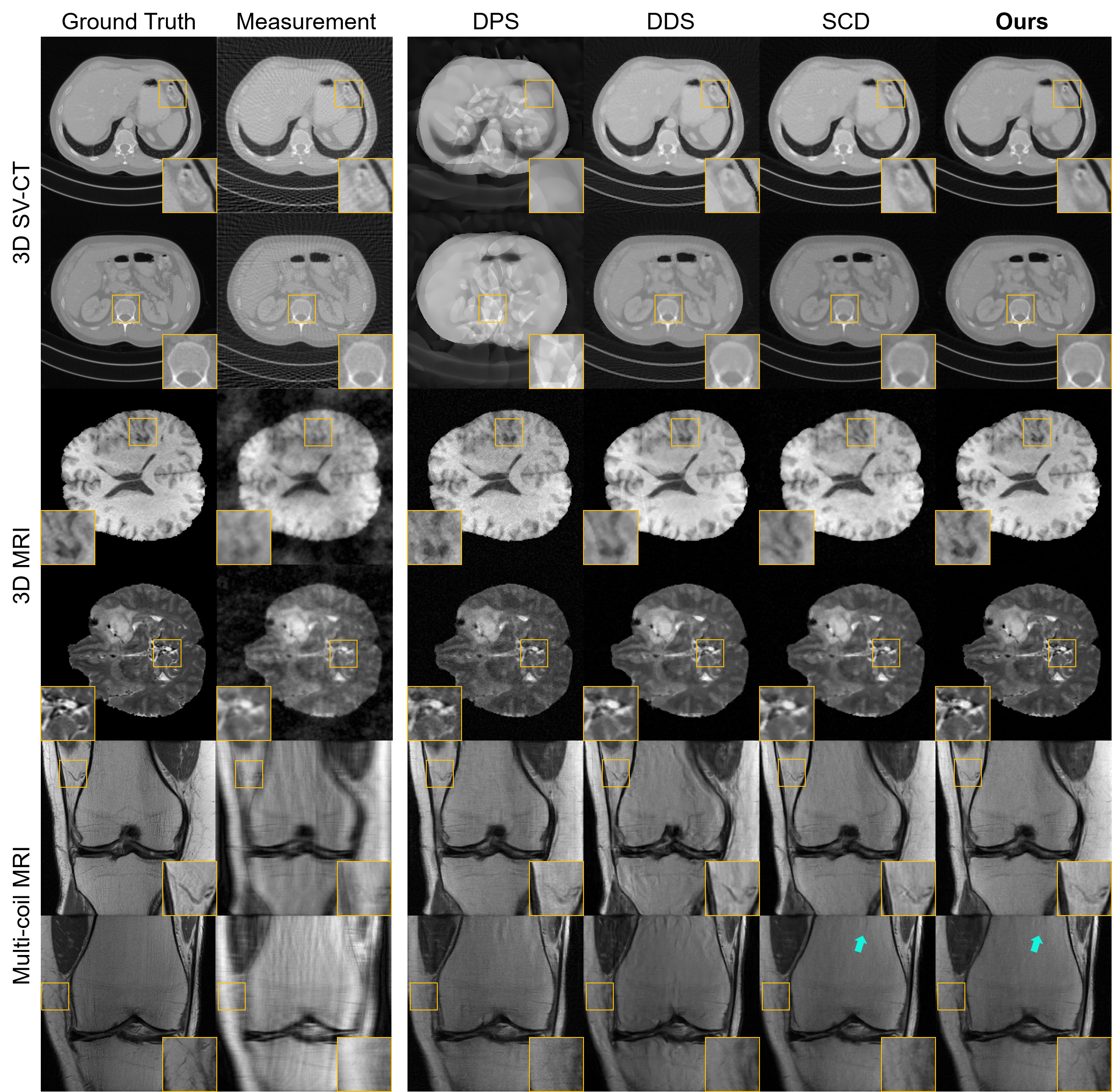}
\caption{Additional results on 3 different tasks. (row 1-2): 3D SV-CT, (row 3-4): 3D MRI, (row 5-6): CS-MRI. Comparison against DPS~\cite{chung2023diffusion}, DDS~\cite{chung2024decomposed}, and SCD~\cite{barbano2023steerable}. \textbf{Ours}: D3IP (base). Cyan arrows indicate regions of remaining artifacts even after adaptation with SCD. Green boxes illustrate the acquisition scheme of the measurement (acquisition angle, sub-sampling pattern).}
\vspace{-0.2cm}
\label{fig:additional_results}
\end{figure}

\begin{figure}[!t]
\centering
\includegraphics[width=\linewidth]{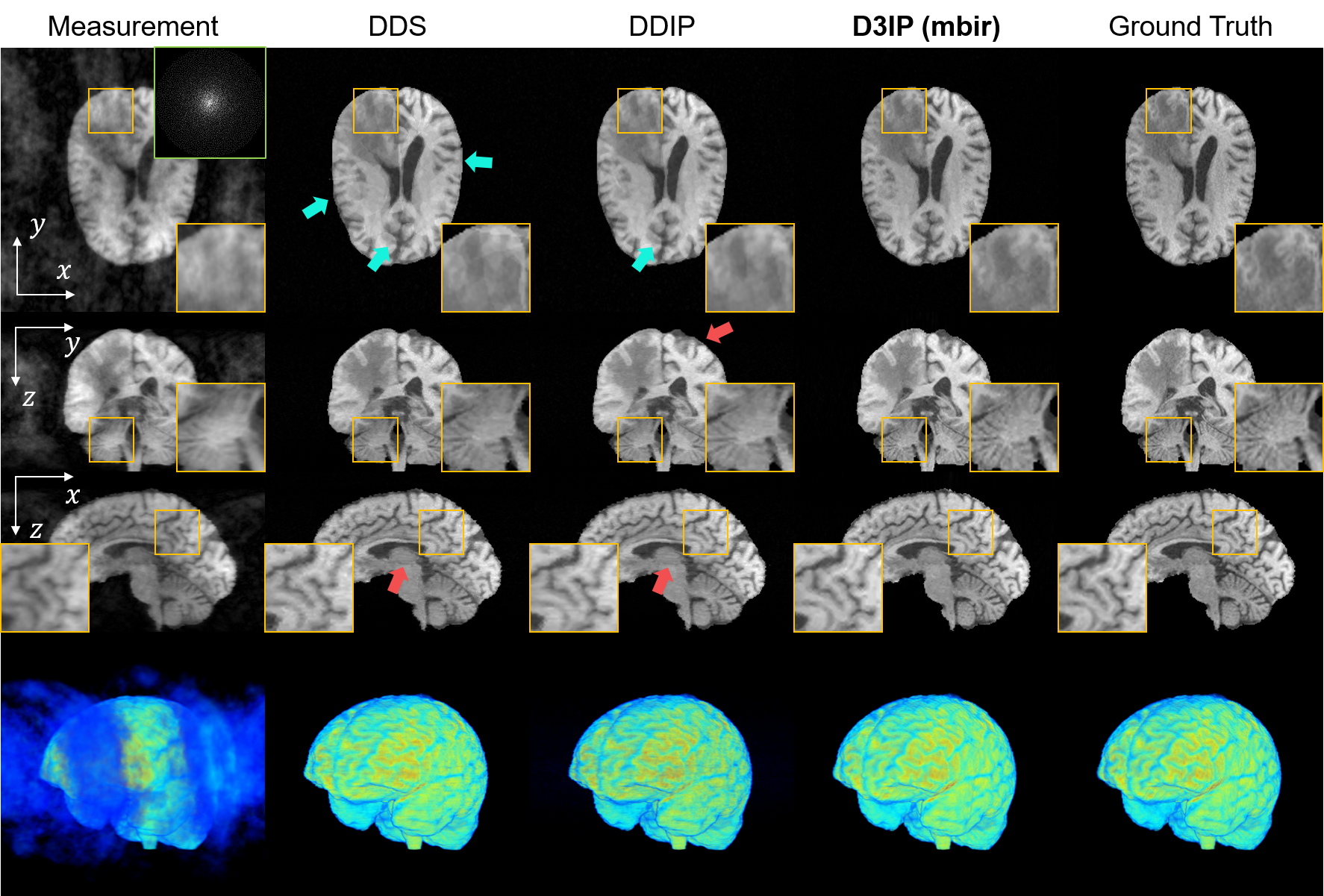}
\caption{3D-MRI reconstruction with DDS~\cite{chung2024decomposed}, DDIP, D3IP (mbir). Cyan and red arrows indicate artifacts from prior mismatch and slice-wise independent reconstruction, respectively. 1-4$^{\rm th}$ row: $xy, yz, xz$ slice, and 3D rendering.}
\vspace{-0.2cm}
\label{fig:additional_results_3d}
\end{figure}

\end{document}